\documentclass[runningheads]{llncs}
\usepackage[T1]{fontenc}
\usepackage{booktabs}
\usepackage{graphicx,verbatim}
\usepackage{placeins}
\usepackage{amsmath}
\usepackage{amssymb}
\usepackage{comment}
\usepackage{float}

\begin{document}
\title{Retina-RAG: Retrieval-Augmented Vision–Language Modeling for Joint Retinal Diagnosis and Clinical Report Generation}
\titlerunning{~}
%

\author{Abdelrahman Zaian\inst{1} \and
        Sheethal Bhat\inst{1} \and
        Mohamed Abdalkader\inst{2} \and
        Andreas Maier\inst{1}}
\authorrunning{~}
\institute{Friedrich-Alexander-Universität, Germany \\
    \email{\{abdelrahman.zaian, sheethal.bhat, andreas.maier\}@fau.de}
    \and
    \email{Mohameed.Abdalkadeer@gmail.com}}

\maketitle              
\begin{abstract} 
Diabetic Retinopathy (DR) is a leading cause of preventable blindness among working-age adults worldwide, yet most automated screening systems are limited to image-level classification and lack clinically structured reporting. 
We propose Retina-RAG, a low-cost modular framework that jointly performs DR severity grading, macular edema (ME) detection, and report generation. The architecture decouples a high-performance retinal classifier and a parameter-efficient vision–langu\-age model (Qwen2.5-VL-7B-Instruct) adapted via Low-Rank Adaptation \- (LoRA), enabling flexible component integration. A retrieval-augmented generation (RAG) module injects curated ophthalmic knowledge together with structured classifier outputs at inference time to improve diagnostic consistency and reduce hallucinations. 
Retina-RAG achieves an F1-score of 0.731 for DR grading and 0.948 for ME detection, substantially outperforming zero-shot Qwen (0.096, 0.732) and MMed-RAG (0.541, 0.641) on a retinal disease detection dataset with captions. For report generation, Retina-RAG attains ROUGE-L 0.438 and SBERT similarity 0.884, exceeding all baselines. The full framework operates on a single consumer-grade GPU, demonstrating that clinically structured retinal AI can be achieved with modest computational resources. Source code at: \url{https://xx/xxx.git}


\end{abstract}

\section{Introduction}

Diabetic retinopathy (DR) is a leading cause of preventable blindness among working-age adults worldwide \cite{Teo2021DR}. According to the Global Burden of Disease Study 2021, the prevalence of vision impairment attributable to DR is projected to rise by 55.6\% to 160.50 million cases by 2045 \cite{Pan2025GBD}. This increase is driven by urbanisation, an ageing population, and the continued expansion of the diabetes epidemic in low- and middle-income countries~\cite{Teo2021DR}. Despite this growing burden, access to trained ophthalmologists is still severely limited in many high-prevalence regions \cite{Pan2025GBD}, creating an urgent need for scalable, automated screening that delivers accurate and clinically interpretable results.

Although deep learning systems have achieved strong performance in automated DR severity grading, most existing approaches remain limited to image-level classification without producing clinically interpretable reports suitable for real-world deployment. CNN-based architectures such as EfficientNet~\cite{Tan2019EfficientNet} and ResNet~\cite{He2016ResNet} variants have demonstrated competitive grading accuracy even under significant class imbalance~\cite{Humayun2023DR,Arora2024Eff}. Hybrid CNN–Transformer and ensemble methods have further advanced classification performance on standard benchmarks~\cite{Xu2024Hybrid}. This limitation persists even at the commercial level: LumineticsCore (formerly IDx-DR), the first FDA-cleared autonomous AI system for DR detection deployed across primary care settings in the United States, outputs only a binary referral decision without any structured clinical description of retinal findings~\cite{Abramoff2018IDxDR}. These systems output discrete severity labels without clinical justification, limiting their translational value in ophthalmic workflows.

Vision–Language models (VLM)s such as LLaVA-Med extend instruction tuning to medical imaging tasks \cite{LLaVAMed2023}. However, systematic evaluations show that both general-purpose and medically adapted VLMs frequently generate factually incorrect outputs in medical settings, raising serious concerns about their reliability for clinical use \cite{MedVH2024}. Dedicated hallucination benchmarks further document this, revealing consistent failure patterns across multiple medical imaging modalities \cite{MedHallMark2024}. Moreover, adapting large-scale VLMs often demands substantial computational resources, limiting deployment in resource-constrained clinical settings \cite{He2024PeFoMed},  motivating the need for parameter-efficient and modular alternatives.

Retrieval-Augmented Generation (RAG) has been proposed to improve factual reliability by conditioning report generation on externally retrieved medical knowledge\cite{Lewis2020RAG}. RULE demonstrates meaningful gains in medical visual question answering through preference-based retrieval alignment \cite{RULE2024}. MMed-RAG further extends this by introducing a domain-aware retrieval mechanism that improves factual accuracy across diverse medical tasks \cite{MMedRAG2024}. However, current RAG frameworks are not designed for DR screening’s diagnostic setup, which combines imbalanced multi-class severity grading with binary Macular Edema (ME) detection.

\begin{figure}[ht]
  \centering
  \includegraphics[width=0.85\textwidth]{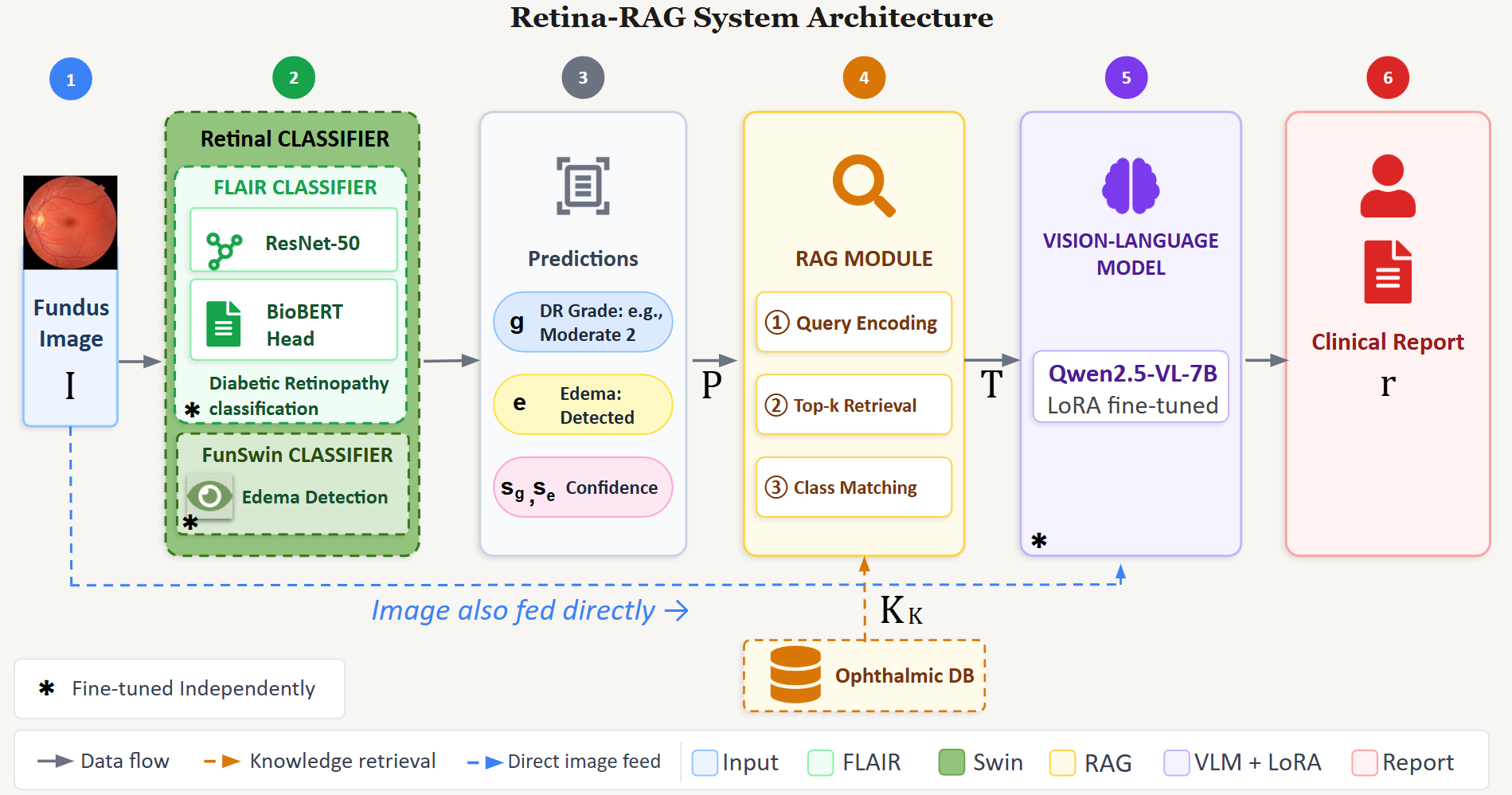}
  \caption{Overview of Retina-RAG for both training and inference. We use a dual-branch retinal classifier to process the fundus image. The classifier predicts either DR severity and ME, producing structured text outputs $\mathcal{P}$. These predictions are serialized to query an ophthalmic knowledge base, retrieving $k$ task-relevant snippets $\mathcal{K}_k$. The fundus image, classifier outputs, and retrieved knowledge are composed into a classifier-guided prompt $\mathcal{T}$ for a LoRA-adapted VLM (Qwen2.5-VL), which generates the final clinical report.}
  \label{fig:architecture}
\end{figure}

To address this gap, we introduce \textbf{Retina-RAG}, a modular framework for joint DR severity grading, ME detection, and structured report generation. Retina-RAG, as seen in Fig.~\ref{fig:architecture} separates a high-performance retinal classifier from a parameter-efficient VL  backbone and integrates classifier-guided retrieval prompting, injecting structured diagnostic predictions alongside curated ophthalmic knowledge at inference. This design aligns language generation with task-specific classification outputs, improving diagnostic consistency while avoiding computationally expensive end-to-end (e2e) multimodal training.

\paragraph{\textbf{Main contributions:}} We
1) propose Retina-RAG, a diagnostic-aware framework that separates DR severity grading and ME detection from report generation, enabling modular optimization while preserving clinical task alignment.
2) introduce a classifier-guided retrieval prompting mechanism that conditions language generation on structured diagnostic predictions and curated ophthalmic knowledge, enforcing consistency between classification and reporting without costly e2e multimodal training.
3) provide comprehensive evaluation on a retinal disease and caption dataset \cite{AbdalkaderRDD2023}, demonstrating improved grading (F1: 0.721), ME detection (F1: 0.948), and report quality (ROUGE-L 0.419; SBERT 0.874) using Retina-RAG in comparison to zero-shot (ZS) and medical RAG baselines.

\section{Materials and methods}

\subsection{Dataset details and preparation}
We evaluate Retina-RAG on Dataset $\mathcal{A}$, the Retinal Disease 
Detection dataset~\cite{AbdalkaderRDD2023}, comprising 2,254 colour 
fundus images (1,577 train / 339 val / 338 test) at $\approx$2,000$\times$1,333 pixels, each annotated with a five-class DR grade and binary ME label. DR grades are severely imbalanced: No~DR dominates at 52.6\%, followed by Moderate (22.7\%), Mild (13.0\%), Severe (7.4\%), and Proliferative (4.3\%). ME labels are 19.6\% positive. Dataset $\mathcal{B}$, MESSIDOR~\cite{Decenciere2014Messidor}, provides 1,200 additional fundus images at 2,240$\times$1,488 with retinopathy grade and ME risk annotations, used solely for supplementary ME classifier training. No patient overlap exists between splits. To address class imbalance, inverse-frequency weighting is applied during training. Images are resized to 512$\times$512 for DR classification, 224$\times$224 for ME classification, and 448$\times$448 for the LoRA-adapted VLM, all normalised with ImageNet statistics ($\mu$\,=\,[0.485, 0.456, 0.406], $\sigma$\,=\,[0.229, 0.224, 0.225]). VLM inputs use LANCZOS resampling; ME classifier inputs apply morphological top-hat and black-hat transforms (15$\times$15 elliptical kernel) on the green channel for lesion contrast enhancement.

\subsection{Methodology}

\textbf{Retina-RAG} is implemented as a modular three-stage pipeline. As illustrated in Fig.~\ref{fig:architecture}, the system comprises: 
(i)~a dual-branch retinal classifier, (ii)~a classifier-guided RAG module, and (iii)~a 
LoRA-adapted VLM for report generation. The classifier and VLM are trained independently, 
enabling low-resource adaptation and flexible component replacement.
The Classifier processes a fundus image $\mathcal{I}$, producing a DR severity grade 
$g\in\{0,\ldots,4\}$, a ME label $e\in\{0,1\}$, with confidence scores $s_g$ and $s_e$ respectively.  
$\mathcal{P}=\{g,e,s_g,s_e\}$ is a structured text instruction to the RAG module to retrieve $k$ task snippets $\mathcal{K}_k$. The VLM processes $\mathcal{T}$ = \{$\mathcal{P}$,  $\mathcal{K}$\}, to generate the final report $r$.


\subsubsection{(i) Retinal classifier} comprises of two independently fine-tuned classifier modules. These are selected based on their performance on DR and ME detection. 

\paragraph{\textbf{DR grading branch:}}
We adopt FLAIR~\cite{FLAIR2025}, an ophthalmology foundation model pretrained on 
288{,}307 fundus images across 101 ophthalmic labels, as the DR grading backbone. 
FLAIR combines a ResNet-50 visual encoder with a BioBERT text head, aligned via 
contrastive pretraining on expert-curated clinical descriptions. We adapt FLAIR for 
five-class DR severity grading 
($g \in \{0,1,2,3,4\}$ corresponding to No~DR, Mild, Moderate, Severe, 
and Proliferative DR, respectively.
using linear probing: the FLAIR backbone is frozen and a single linear classification 
head is trained with a standard CE loss. 

\paragraph{\textbf{ME branch:}}
Binary ME detection is processed by an ImageNet-pretrained FunSwin Transformer 
classifier~\cite{Liu2021SwinTransformer}, fine-tuned on our 
training set using a standard BCE loss. 
Decoupling the two branches preserves the domain-specific representations of each 
pretrained backbone and allows independent optimization per task. 
At inference, both branches run in parallel and their outputs are serialised into $\mathcal{P} = \{g,\, e,\, s_g,\, s_e\}$, which is passed downstream to both the RAG module and the VLM prompt.

\subsubsection{(ii) RAG module}
grounds report generation from structured classifier outputs and retrieves
ophthalmic knowledge. This reduces hallucinations and enforces consistency between predicted 
grades and generated text~\cite{RULE2024}.

\paragraph{\textbf{Knowledge Base Construction:}}
An offline knowledge base of curated DR clinical report descriptions, 
grounded in established severity criteria~\cite{Wilkinson2003ICDR,AAO2019PPP}, 
is embedded with \texttt{MedEmbed-small-v0.1} and indexed via 
FAISS~\cite{johnson2019billion} for nearest-neighbour retrieval.

\paragraph{\textbf{Structured Query Formation and Class Matching:}}
At inference, $\mathcal{P}$ is serialized from natural-language (e.g.,~\textit{``Moderate DR, ME detected, confidence 0.87''}) to query embeddings $\mathbf{q} \in \mathbb{R}^d$. We perform \emph{class matching} based on the grades $g$ and $e$, thereby ensuring retrieved snippets are semantically aligned with the predicted diagnostic class, 
rather than with superficial visual features of the query image.


\paragraph{\textbf{Retrieval and prompt construction:}}
The top-$k$ ($k{=}3$) knowledge snippets 
$\mathcal{K}_k = \{d_1, \ldots, d_k\}$ are retrieved by cosine similarity between $\mathbf{q}$ and  $\mathbf{d}_i \in \mathbb{R}^d$, the embedding of the $i$-th knowledge entry.
The fundus image $\mathcal{I}$, classifier outputs $\mathcal{P}$, and retrieved snippets 
$\mathcal{K}_k$ are composed into a structured  $\mathcal{T}$:
$\texttt{Prompt}\!\left(\mathcal{I},\,\mathcal{P},\,\mathcal{K}_k\right).$
The VLM is instructed to treat $\mathcal{P}$ as the primary diagnostic anchor and 
$\mathcal{K}_k$ as supporting clinical context as below,
\begin{quote}
\small\textit{``You are an expert ophthalmologist. The classifier predicts: 
$\{\mathcal{P}\}$. Relevant clinical context: $\{\mathcal{K}_k\}$. Generate a concise 
diagnostic report for this fundus image.''}
\end{quote}

\subsubsection{(iii) LoRA-Adapted VLM}
Report generation uses Qwen2.5-VL-7B-Instruct~\cite{Qwen2VL}, chosen for its strong instruction-following performance under resource-constrained deployment~\cite{He2024PeFoMed}. Fine-tuning is performed using Unsloth's 
\texttt{FastVisionModel}~\cite{unsloth2024} with LoRA~\cite{Hu2022LoRA}, applied to attention and MLP projections in both the vision 
and language blocks. The adapted weight update is:
\begin{equation}
    W' = W + \Delta W = W + \tfrac{\alpha}{r}BA,
\end{equation}
where $W$ is frozen, and $A \in \mathbb{R}^{r \times d}$, $B \in \mathbb{R}^{d \times r}$ 
are trainable low-rank matrices with $r \ll d$. The model is trained via supervised 
fine-tuning (SFT) \cite{Ouyang2022InstructGPT,unsloth2024} using causal language modelling loss on the assistant response tokens $r$, where $r_t$ is the $t$-th token of the reference report and $\mathcal{T}$ is the structured prompt.

\section{Experiments}

\subsubsection{Evaluation protocol:}
Classification performance under class imbalance is evaluated using 
F1-score, precision, and recall. Accuracy is not reported due to its 
sensitivity to skewed label distributions. Report generation quality 
is assessed using BLEU-4~\cite{Papineni2002BLEU}, ROUGE-1 and ROUGE-L~\cite{Lin2004ROUGE}, and semantic similarity computed with SBERT~\cite{Reimers2019SBERT} and ClinicalBERT~\cite{Huang2019ClinicalBERT} embeddings.

\subsubsection{Implementation details:}

FLAIR is fine-tuned for 50 epochs using Adam (LR $1\times10^{-4}$ for encoder–decoder, $1\times10^{-5}$ for visual extractor, weight decay $1\times10^{-5}$)~\cite{Hu2022LoRA}, while FunSwin uses AdamW (LR $1\times10^{-4}$)~\cite{Liu2021SwinTransformer}. Qwen2.5-VL-7B-Instruct is adapted via LoRA ($r{=}64$, $\alpha{=}128$) applied to attention and MLP layers in both vision and language branches; the 4-bit base model is frozen and only LoRA parameters are updated. Training employs 8-bit AdamW with cosine decay (warmup 0.15), weight decay 0.01, batch size 2 with gradient accumulation (×4), and gradient clipping 0.5; loss is computed only over assistant-response tokens (SFT). Minority DR grades are upsampled with $\pm15^\circ$ rotation and brightness augmentation. The RAG module retrieves $k{=}3$ snippets via FAISS using \texttt{MedEmbed-small-v0.1}. Experiments run on a single RTX 1650 (4GB), and results report the mean of three seeds.

\paragraph{\textbf{Baselines:}}
We compare Retina-RAG against (1) ZS Qwen, the unmodified Qwen2.5-VL-7B-Instruct prompted with the fundus image and a generic ophthalmology instruction, without fine-tuning or retrieval; and (2) MMed-RAG~\cite{MMedRAG2024}, a state-of-the-art (SOTA) as well as FFA-IR \cite{FFAIR2021}. 

\subsection{Results}

Table~\ref{tab:classification} shows that Retina-RAG achieves the best overall classification performance, with macro-F1 of 0.731 for DR severity and 0.948 for ME. For DR grading, Retina-RAG substantially outperforms ZS Qwen (0.096) and MMed-RAG (0.541), and improves over FFA-IR (0.660), while remaining competitive in AUC (0.774 vs. 0.837). For ME detection, Retina-RAG achieves the highest F1 (0.948), slightly surpassing FFA-IR (0.947), and clearly outperforming ZS Qwen (0.732) and MMed-RAG (0.641). These results highlight the importance of domain adaptation and retrieval grounding for fine-grained retinal classification.

Table~\ref{tab:generation} compares report generation quality. Retina-RAG achieves a ROUGE-L of 0.419 and SBERT similarity of 0.874, substantially outperforming MMed-RAG (0.353/0.786) and the ZS baseline (0.163/0.764). ClinicalBERT similarity (0.972) further confirms strong clinical coherence and semantic alignment with reference reports. While FFA-IR shows competitive classification performance, it is limited to short reports ($\leq$60 words), whereas Retina-RAG supports long-form generation (approx. 1,000 words) while maintaining high diagnostic and linguistic quality.

\begin{table}[t]
\caption{Classification comparison between Retina-RAG and various SOTA methods on RDD test dataset. Precision, Recall, F1 are weighted-averaged and AUC is macro-averaged across DR categories. Results are mean of 3 runs.}\label{tab:classification}
\centering
\fontsize{8}{10}\selectfont
\setlength{\tabcolsep}{4.8pt}
{%
\begin{tabular}{l|cccc|cccc}
\toprule
& \multicolumn{4}{c|}{\textbf{DR Severity (5-class)}}
& \multicolumn{4}{c}{\textbf{ME (Binary)}} \\
\textbf{Method} 
  & \textbf{Prec.} & \textbf{Rec.} & \textbf{F1} & \textbf{AUC}
  & \textbf{Prec.} & \textbf{Rec.} & \textbf{F1} & \textbf{AUC} \\
\midrule
ZS Qwen \cite{Qwen25VL2025}
  & 0.076 & 0.204 & 0.096 & 0.511
  & 0.743 & 0.722 & 0.732 & 0.598 \\
MMed-RAG \cite{MMedRAG2024}
  & 0.679 & 0.450 & 0.541 & 0.603
  & 0.682 & 0.621 & 0.641 & 0.494 \\
FFA-IR \cite{FFAIR2021}
  & 0.668 & 0.650 & 0.660 & \textbf{0.837} 
  & 0.942 & 0.945 & 0.947 & \textbf{0.934} \\
{Retina-RAG (Ours)}
  & \textbf{0.778} & \textbf{0.795} & \textbf{0.731} & 0.774
  & \textbf{0.949} & \textbf{0.947} & \textbf{0.948} & 0.897 \\
\bottomrule
\end{tabular}}
\end{table}

\begin{table}[t]
\caption{Report generation quality on the RDD test set denoted by NLP metrics for Retina-RAG and other SOTA methods.}\label{tab:generation}
\centering
\fontsize{8}{10}\selectfont
\begin{tabular}{lccccc}
\toprule
\textbf{Method} & \textbf{BLEU-4} & \textbf{ROUGE-1} & \textbf{ROUGE-L} & \textbf{SBERT} & \textbf{Clin-BERT} \\
& \cite{Papineni2002BLEU} & \cite{Lin2004ROUGE} & \cite{Lin2004ROUGE} & \cite{Reimers2019SBERT} & \cite{Huang2019ClinicalBERT} \\
\midrule
ZS Qwen \cite{Qwen25VL2025}           & 0.014 & 0.175 & 0.163 & 0.764 & 0.921 \\
MMed-RAG   \cite{MMedRAG2024}               & 0.082 & 0.365 & 0.353 & 0.786 & 0.923 \\
FFA-IR  \cite{FFAIR2021}              & \textbf{0.235} & \textbf{0.529} & 0.383 & 0.869 & \textbf{0.982} \\
{Retina-RAG (Ours)} & 0.194 & 0.461 & \textbf{0.438} & \textbf{0.884} & 0.975 \\
\bottomrule
\end{tabular}
\end{table}


\paragraph{\textbf{Qualitative Results:}}

\begin{figure}[t]
\centering
\includegraphics[width=1.0 \linewidth]{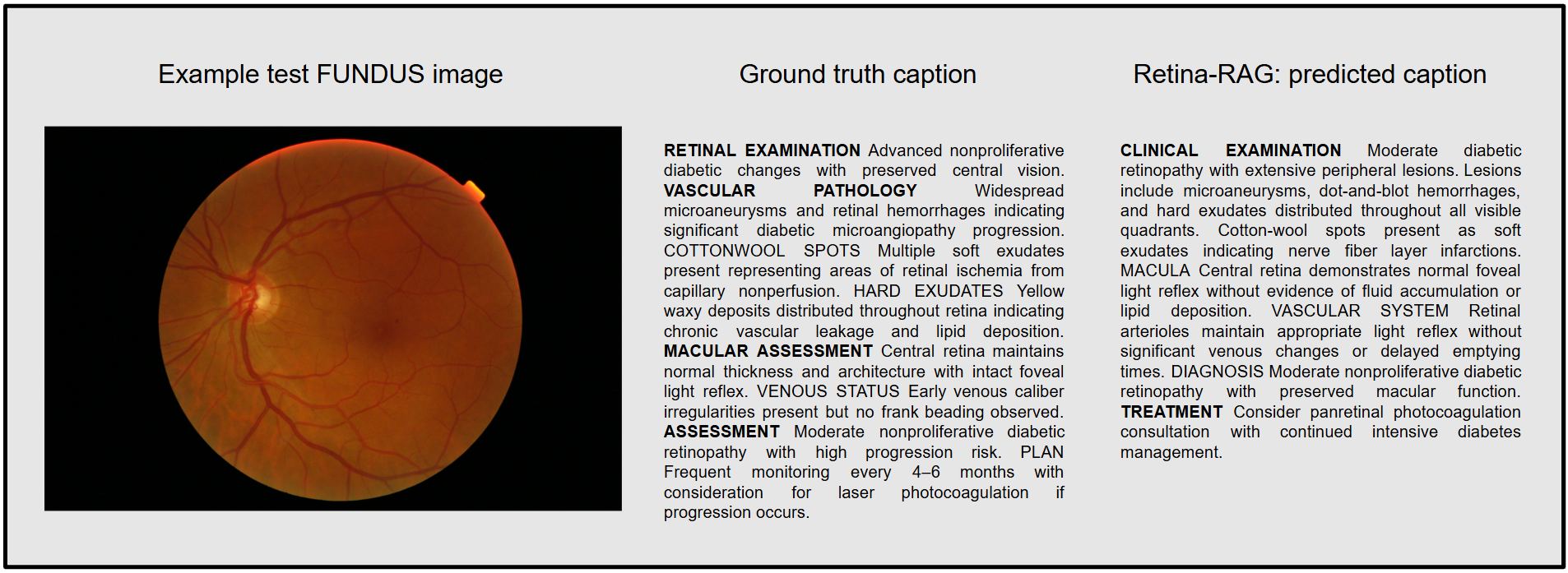}
\caption{Qualitative example from RDD test dataset (Image ID 00804) with groundtruth and Retina-RAG predicted report. Image indicates DR Grade 2 (Moderate); ME Risk 0 (No Risk), which is reflected in the predicted report.}
\label{fig:00804}
\end{figure}

Fig.~\ref{fig:00804} denotes an example of a fundus image from the RDD test set with the ground truth and predicted caption. This example shows how Retina-RAG leverages retrieved Grade 2 clinical context to ground the predicted caption, correctly identifying microaneurysms, cotton-wool spots, and dot-and-blot hemorrhages consistent with Moderate Non-Proliferative DR. The macular assessment correctly indicates no signs of diabetic ME, and no hallucinated severe features present in the output.

\paragraph{\textbf{Cost impact:}}

Retina-RAG is designed for resource-constrained clinical deployment. The dual-branch classifiers are fine-tuned via linear probing on either single GPU or Google Collab TPU. Qwen2.5-VL-7B is adapted via LoRA (Unsloth), training only 1.23\% of parameters (103M/8.4B) in 367 minutes with 10.5GB peak memory. The full pipeline runs on a single GPU at $<\$0.0002$ per report, ~\cite{hyperbolic2024}, and to the best of our knowledge, orders of magnitude cheaper than proprietary models $(\sim \$0.01 - \$0.05$ per query), supporting scalable screening in low-resource settings.

\paragraph{\textbf{Ablation Studies:}}

We ablate the system into five components: (1) frozen VLM baseline, (2) fine-tuned VLM only, (3) classifier $+$ frozen VLM, (4) classifier $+$ fine-tuned VLM, and (5) the full Retina-RAG model with retrieval. Adding the classifier substantially improves DR grading performance, fine-tuning the VLM improves report quality, and incorporating retrieval provides the final gain in diagnostic consistency. We evaluate additional classifier in place of FLAIR, including a single VGG-16 \cite{Rocha2022VGG16DR} and a dual GoogleNet  $+$ ResNet-18 \cite{Butt2022HybridDR} configuration with an RBF-SVM \cite{Behera2020RBFSVM} layer at the end for classification. The GoogleNet and ResNet-18 are trained similar to the Retinal classifier.

\begin{table}[t]
\caption{Component-wise ablation on DR grading and edema 
detection. ZS= Zero shot, FT=Fine-tuned.}\label{tab:ablation_cls}
\centering
\fontsize{8}{10}\selectfont
\setlength{\tabcolsep}{5pt}
{%
\begin{tabular}{llcccccc}
\toprule
\textbf{Config} & \textbf{Classifier} &
\multicolumn{3}{c}{\textbf{DR Severity (5-class)}} &
\multicolumn{3}{c}{\textbf{ME (Binary)}} \\
\cline{3-8}
& & \textbf{Prec.} & \textbf{Rec.} & \textbf{F1} &
    \textbf{Prec.} & \textbf{Rec.} & \textbf{F1} \\
\midrule
ZS Qwen (Frozen VLM)  & --  & 0.076 & 0.204 & 0.096 
  & 0.743 & 0.722 & 0.732  \\
FT VLM              & --      & 0.395 & 0.252 & 0.160 & 0.731 & 0.479 & 0.525 \\
\midrule
Classifier + ZS Qwen       & FLAIR   & 0.667 & 0.462 & 0.470 & 0.910 & 0.911 & 0.910 \\
                         & VGG16   & 0.537 & 0.396 & 0.411 & 0.941 & 0.941 & 0.938 \\
                         & RBF-SVM & 0.691 & 0.459 & 0.470 & 0.899 & 0.899 & 0.899 \\
\midrule
Classifier + FT VLM       & FLAIR   & 0.782 & 0.752 & 0.698 & 0.881 & 0.876 & 0.878 \\
                         & VGG16   & 0.444 & 0.568 & 0.487 & 0.850 & 0.861 & 0.846 \\
                         & RBF-SVM & 0.682 & 0.701 & 0.674 & 0.860 & 0.861 & 0.861 \\
\midrule
Retina-RAG (Ours) & FLAIR   & \textbf{0.778} & \textbf{0.795} & \textbf{0.731}  & \textbf{0.949} & \textbf{0.947} & \textbf{0.948}\\
\bottomrule
\end{tabular}}
\end{table}

\begin{table}[t]
\caption{Component-wise ablation on report generation using standard NLP metrics. ZS=Zero shot, FT=Fine-tuned.}\label{tab:ablation_gen}
\centering
\fontsize{8}{10}\selectfont
{%
\begin{tabular}{llccccc}
\toprule
\textbf{Config} & \textbf{Classifier} &
\textbf{BLEU-4} & \textbf{ROUGE-1} & \textbf{ROUGE-L} &
\textbf{SBERT} & \textbf{Clin-BERT} \\
& & \cite{Papineni2002BLEU} & \cite{Lin2004ROUGE} & \cite{Lin2004ROUGE} & \cite{Reimers2019SBERT} & \cite{Huang2019ClinicalBERT} \\
\midrule
ZS Qwen (Frozen VLM)         & -- & 0.014 & 0.175 & 0.163 & 0.764 & 0.921 \\
FT VLM & --      & 0.114 & 0.342 & 0.333 & 0.840 & 0.960 \\
\midrule
Classifier + ZS Qwen       & FLAIR   & 0.016 & 0.166 & 0.158 & 0.695 & 0.895 \\
                         & VGG16   & 0.019 & 0.176 & 0.167 & 0.699 & 0.898 \\
                         & RBF-SVM & 0.016 & 0.170 & 0.162 & 0.700 & 0.896 \\
\midrule
Classifier + FT VLM       & FLAIR   & 0.186 & 0.439 & 0.422 & 0.881 & 0.975 \\
                         & VGG16   & 0.165 & 0.410 & 0.395 & 0.863 & 0.970 \\
                         & RBF-SVM & 0.186 & 0.419 & 0.412 & 0.871 & 0.971 \\
\midrule
Retina-RAG (Ours) & FLAIR & \textbf{0.194} & \textbf{0.461} & \textbf{0.438} & \textbf{0.884} & \textbf{0.975} \\
\bottomrule
\end{tabular}}
\end{table}

\section{Conclusion}

We present Retina-RAG, a modular framework for joint retinal diagnosis and long-form report generation, demonstrating strong performance at low computational cost at $<$ 2.23\% parameter tuning. 
A primary limitation is that evaluation is conducted on a single benchmark dataset, which may not fully reflect variability across imaging devices, patient populations, and grading protocols. Future work will investigate cross-dataset generalization and prospective clinical validation.
\FloatBarrier

\newpage
%
%
%
%

\end{document}